# IBERT: Idiom Cloze-style reading comprehension with Attention

Ruiyang Qin*
rqin37@gatech.edu
Georgia Institute of technology
Atlanta, Georgia, USA

Haozheng Luo*
hluo76@gatech.edu
Georgia Institute of technology
Atlanta, Georgia, USA

Zheheng Fan*
zfan71@gatech.edu
Georgia Institute of technology
Atlanta, Georgia, USA

Ziang Ren*
ziang_ren@gatech.edu
Georgia Institute of technology
Atlanta, Georgia, USA

## ABSTRACT
Idioms are special fixed phrases usually derived from stories. They are commonly used in casual conversations and literary writings. Their meanings are usually highly non-compositional. The idiom cloze task is a challenge problem in Natural Language Processing (NLP) research problem. Previous approaches to this task are built on sequence-to-sequence (Seq2Seq) models and achieved reasonably well performance on existing datasets. However, they fall short in understanding the highly non-compositional meaning of idiomatic expressions. They also do not consider both the local and global context at the same time. In this paper, we proposed a BERT-based embedding Seq2Seq model that encodes idiomatic expressions and considers them in both global and local context. Our model uses XLNET as the encoder and RoBERTa for choosing the most probable idiom for a given context. Experiments on the EPIE Static Corpus dataset show that our model performs better than existing state-of-the-arts.

## 1 INTRODUCTION
The cloze test is a test in which the participant is asked to supply words that have been removed from a passage. Cloze tests are generally used to evaluate a participant's ability to comprehend the given text. One significant difference between the cloze task from other Natural Language Processing (NLP) problems is that the former requires a considerably larger long-term memory to make decisions and the ability to comprehend this content. Solving the cloze tests will provide insight into existing NLP tasks in text comprehension. In this paper, we employ a BERT-based sequence-to-sequence (Seq2Seq) model to solve the cloze task. Given a specific context in the form of a passage, the solution of a cloze problem based on this context is generated in two steps: understanding the meaning of the idiom and choosing the correct idiom for each "blank" – where the original word is removed – in the passage. Previous research approaches NLP problems similar to the cloze task by applying neural network models such as the Knowledgeable Reader [9] and Entity Tracking [5]. These aforementioned approaches treat idioms as regular phrases. However, idiomatic expressions usually have meanings that are highly non-compositional and should not be understood from a literal standpoint. For example, if we look at the idiom word by word literally, the expression "it's raining cats and dogs" describes cats and dogs falling form the sky. This understanding is apparently incorrect as we generally use this expression to describe a heavy rain. Failure to understand the correct meaning of idiomatic expressions is detrimental to the decision-making process of the models, as even if a model is good enough to comprehend the passage itself, it cannot associate an expression that, from the literal perspective, is completely unrelated to the context. This calls for different approaches to the cloze task as our model is required to not only provide a grammatically sound candidate, but also ensure that the semantical meanings are coherent as well. Previous work on contextual relation understanding has yielded solutions with excellent performance. The BERT [3] performs well at understanding the contextual meaning of a given context. Pretrained BERT and other similar models can be adapted to perform contextual comprehension in other tasks. In this paper, we use pretrained models to solve the context comprehension step of the cloze task. To understand the meaning of idiomatic expressions correctly, we employ another pretrained model XLNET [20] and fine-tune it to solve the cloze task. We train XLNET to learn not only the local context of the removed word (the sentence from which an original word is removed), but also the global context (the entire passage). This knowledge is represented by the last layer output of the hidden layers in XLNET. The contextual embeddings generated by our BERT-based pretrained models are combined with the idiom embeddings from the XLNET model. Softmax [8] is applied to decide on the best word from the combined embeddings. Our models are fine-tuned on the EPIE [14] Static Corpus dataset. The remainder of the paper is organized as follows. Section 2 introduces some related work in cloze-style reading comprehension and modern NLP and Deep Learning technologies. Section 3 describes the method we developed to solve the challenge our problem. Section 4 describes our experiment to examine the performance of the models. Finally, Section 5 and 6 present the result and conclusion of our experiment and research work.

## 2 RELATED WORK
### 2.1 Cloze-style reading comprehension
Cloze-style reading comprehension uses a passage of word tokens $x_{1:n}$, with one token $x_j$ masked; the task is to fill in the masked word y, which was originally at position j. There already have a lot of works achieve a great performance in the cloze work [5, 9, 15].

*All authors contributed equally to this research.

And Researchers created many large-scale cloze-style reading comprehension datasets yet, such as RACE [6], Children's Book Test (CBT)[4]. However, these research work only works on the normal words to finished the Cloze-style reading comprehension, and the idiom phrases are oftentimes non-contextual in the paragraphs. In this research effort, we want to utilize the most advanced technologies in our tasks to get the state of the art performance in the Idiom Cloze-style reading comprehension. The dataset we use EPIE used in this paper is also a large scale cloze-style dataset but focuses on English idiom prediction.

## 2.2 Pre-trained Language Models

During daily usage, enormous sources of unilateral context can influence the model's accuracy. In practice, it's highly possible to have the issue that they do not have symbols after sentences and polysemous condition in the sentence. There are previous researches on improving word embedding [10, 12], but it still cannot help us solve the challenge in our tasks. With the milestone of the appearance of transformer [19], there were several pre-trained models proposed in the NLP area like BERT and XLNet. With the various research, Language model pre-training has been proven to be effective over a list of natural language tasks at both sentence level [2] and token level [18]

## 2.3 Idiom detection

A lot of research work are beginning forces on the idiom cloze-style reading comprehension. The first popular neural network reading comprehension models were the Dual Embedding Model with bert [17]. In this previous work from Tan et al. [17], they use the BERT model to encode the contextual sentence as well as candidate phrases. However, the normal BERT model have the shortcoming to handle the long text sequence and lack of confidence in commonsense pragmatic inference. Also, the BERT model show mediocre performance in negation situation in the sentence. For some Cloze-style reading comprehension, it consist a huge context size and the related keys of cloze choice are far from the masks. Also, we could not ignore the condition of negation and commonsense in the challenge of the Cloze-style reading comprehension.

## 3 METHOD

### 3.1 Task Definition

The main idea about our project is the idiom cloze test. We choose idiom as candidate and take off the candidate from each sentence. For those candidates, they need to be filled into each blank spot. The first thing is to padding each candidate. Then we need to find the most appropriate candidate each time we hit the blank spot.

### 3.2 Padding Sequence

In this part, we begin with processing the candidates. In our project, the first step is to grasp idioms or candidates. We will search for the most appropriate candidate in the following steps. To do search, we have to process those candidates.

The reason we have to do that is because every candidate has different length and it's hard to make any operations on those different-length candidates The default padding and probably the most usual padding is zero-padding, in which we add numbers of zeros in the back of each candidate. For the padding length, we will take the value of the longest candidate and use it as our padding length. This task can be achieved by calling specific functions. We will need to search for longest candidate first.

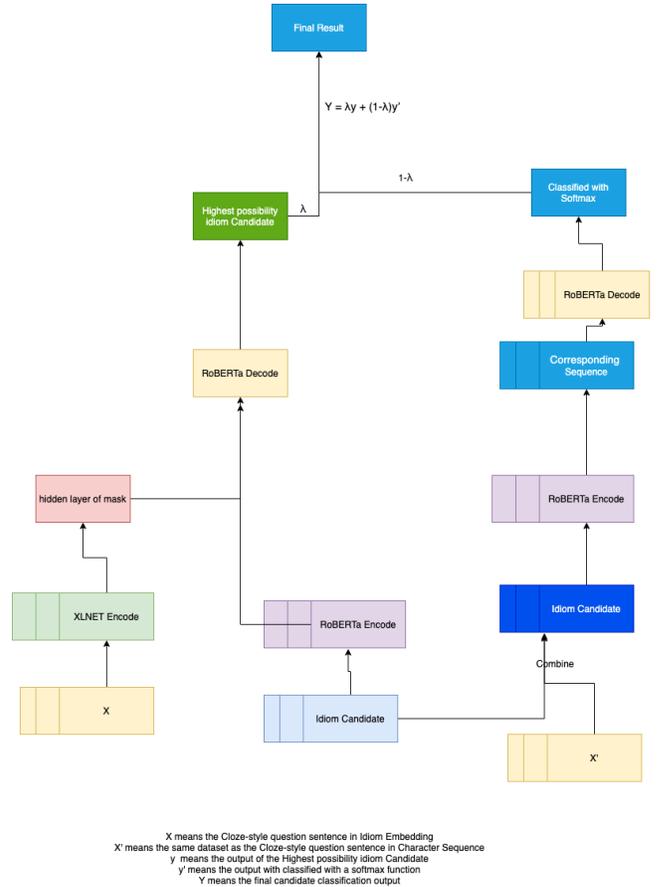

X means the Cloze-style question sentence in Idiom Embedding
X' means the same dataset as the Cloze-style question sentence in Character Sequence
y means the output of the Highest possibility idiom Candidate
y' means the output with classified with a softmax function
Y means the final candidate classification output

**Figure 1: Our Method Diagram about how our Attention Baselines works**

### 3.3 Attention Baselines

Previous methods applied to the BERT-based Dual Embedding [17] are only based on the BERT architecture. Because of the shortcoming in BERT and success of XLNET for many NLP task, especially reading comprehension, we proposed to present new methods (shown as Figure 1) based on XLNET and BERT to resolve the Idiom Cloze-style reading comprehension. The first one, we treats a English idiom as a sequence of characters and we used the BERT to embedding the passage to analysis the original relationship in each contextual words. We combine the passges with each candidate idiom into a sequence and processes with multiple sequences, one for each candidate. The second baseline we treat the each idiom as a single token which has its own embedding vector and use XLNET to process the passage and then matches the encoded passage with each candidate idiom's embedding.

**Attention Baseline with Idioms as Character Sequence** Attention baseline with idiom as candidate sequence is work to apply the BERT model for idiom cloze-style reading comprehension. Given a passage P = $(p_1, p_2, p_3, \cdots, [MASK], \cdots, p_n)$ and a candidate $d_k \in D$, we first concatenate them into a single sequence ([CLS], $p_1, p_2, p_3, \cdots, d_{n_1}, d_{n_2}, \cdots, d_{n_k}, \cdots, p_n$,[SEQ]), where $d_{n_1}$ to $d_{n_k}$ are the characters and padding of the idiom $d_n$. We can directly use the BERT to process this sequence and obtain the hidden representation for [CLS] in the last hidden layer, denoted by $h_{k,0}^L \in \mathbb{R}^d$. In order to detect the candidate idiom $d_k$ among all the candidate, we use the linear layer to process $h_{k,0}^L$ for $k = 1, 2, \cdots, K$ and use the softmax function with the each probability value of the candidates in D. After that, we will choose the best one candidate as the final one for our cloze-style reading comprehension.

**Attention Baseline with Idiom Embedding** Many idiom are non-compositional and therefore their meaning should not be directly derived from the its full individual characters. For example, "It's a piece of cake" literally means it is a number of cake, but it usually used to describe the task is pretty easy. Therefore, if we only embedding with its each characters meaning will cause a lot of confusion and a single embedding vector for the entire idiom can help the model understands the contextual relationship in the reading comprehension.

In this baseline, instead of concatenating the passage and a candidate answer into a single sequence of BERT model, we separate them. We used the XLNET to process the passage sequence ([CLS], $p_1, p_2, p_3, \cdots, [MASK], \cdots, p_n$,[SEP]). After that, we use the hidden representation of [MASK] at the decode layer, denoted as $h_b^L$, to match each candidate answer with BERT. In this way, no matter how much candidate we are, we only embedding the whole passage once with XLNET model. It can help our work prevent the problem of idiom's non-composition.

We use $d_k$ to denote the embedding vector for candidate $d_k \in D$, the hidden representation $h_b^L$ is work with each candidate idiom via element-wise multiplication. Then the probability of selecting $d_k$ among all the candidate is defined as equation 1.

$$p_k = \frac{exp(w \cdot (d_k \otimes h_b^L) + b)}{\sum_{d'=1}^{D} exp(w \cdot (d_k \otimes h_b^L) + b)} \quad (1)$$

where $w \in \mathbb{R}^d$ and $b \in \mathbb{R}$ are model parameters, and $\otimes$ is element-wise multiplication. To train the model, we use cross entropy loss as the loss function.

### 3.4 Context-aware Pooling

The attention model baseline have potential problem. We can easily observe the idiom are always non-compositional. As a result, in order the idiom candidate can be fit in the passage well, it not only the grammar should fit with the contextual surrounding, also its meaning should also greatly fit the whole passages. As those reason, how to get the candidate not just reasonable in grammar, such as verbs or noun, is importance for us to solve the cloze-style reading comprehension. That is why we need do some development on our baseline in order to context-aware.

As the milestone of the transformer, the more and more pre-train model support the function of the context-aware with the contextual surrounding. Also, some models support the global context-aware, like the BERT work in SQuAD dataset [13]. As a result, we decide to evaluate whether a idiom candidate is suitable in a passage, we need not only understand its neighbour contextual details, also need to know the semantic meaning in the entire passage. That is why we need not only match the idiom candidate to place its context place on, also we must match it meaning with the entire passage. Recall that $H^L = (h_0^L, h_1^L, h_2^L, \cdots, h_n^L)$ represent the hidden states of the last hidden layer of baseline after it process the sequence. Our model with context-aware pooling can be work as equation 2.

$$p_k = \frac{exp(d_k \cdot h_b^L + max_{i=0}^n(d_k \cdot h_i^L))}{\sum_{d'=1}^{D} exp(d_k \cdot h_b^L + max_{i=0}^n(d_k \cdot h_i^L))} \quad (2)$$

### 3.5 Dual Pretrain Attention Model

In our method (Figure 1), we proposed use the linear interpolations [22] to make our solution of idiom detection for the cloze-style reading comprehension become more smooth. We perform linear interpolations in final texture hidden space between both training output, one from the output of classified with the softmax function in Attention Baseline with Idioms as Character Sequence, and the other is the possibility idiom candidate from the Attention Baseline with Idiom Embedding. We use the $\lambda$ as a parameter of the weight to make sure both baseline can work smoothly on the final output. In our model, The $\lambda$ is a sample from beta distribution. We ensure the lambda larger than 0.7, ensure the Attention Baseline with Idioms as Character Sequence dominate the combination.

With the linear interpolation, we can easily get a new possibility of the candidate and we choose the highest one as our final candidate of the cloze-style reading comprehension.

## 4 DATA

We evaluate the accuracy of the classification model with one standard dataset - EPIE consisting of 359 kinds of candidate (listed as table 1), and we make a program to create the classification what idiom should be used in cloze-type reading comprehension sentences. In our program we developed program to use self-attention for the task. After that, We use the training, validation, and test splits as defined in Lee and Dernoncourt [7] to analyse our loss score of the classification.

Table 1 shows the statistics for dataset. There are many number of sentences to classify what kind of the idiom are the most reasonable in the sentence. All of idiom are normally used in the English speaking society and all types of the sentence are the condition of the idioms using.

As table 1 shows, we have 21890 candidate as total. For each candidate, we have a corresponding tag (shown as 2). Both O, B-IDIOM, and I-IDIOM represent the one position of words or special symbol. Given 1 sentence, we need to take off the part which start from the "B-IDIOM" and end at the last position of "I-IDIOM". We got the 21890 sentences after deleting the candidate from the original sentences.

We use 3 types of labels for our data. For all 21890 candidate, we add a ['CLS'] at the beginning of the candidate and add ['SEP'] at end of the candidate in order to label our data. For all sentences which delete the candidate, we add a ['CLS'] at the beginning add ['SEP'] at end in order to label our deleting sentences, in addition to that, we add a ['UNK'] label at the position of the deleting candidate in the 21890 sentences. By the above labeling, we could encode all 21891 candidate and deleting sentences. The example is showing at table 3.

Table 1: Number of Sentences in the Dataset

| Dataset | Train | Validation | Test | T | N |
|---|---|---|---|---|---|
| EPIE | 15k | 5k | 2k | 359 | 21890 |

|T| represents the number of idiom candidate and |N| represents the sentence data size

Table 2: Type of Tags in the Dataset

| Type1 | Type2 | Type3 |
|---|---|---|
| O | B-IDIOM | I-IDIOM |

Table 3: Example of Encoding and Labeling

| |
|---|
| Original Sentence:Anyway , thanks MKM and keep up the good work ! |
| Candidate:keep up the good work |
| Label Candidate:'[CLS]', 'keep', 'up', 'the', 'good', 'work', '[SEP]' |
| encode:11815, 17, 19, 3466, 414, 536, 692, 21, 435, 76, 18, 195, 154, 17, 136, 4, 3 |
| deleting sentence:'Anyway', ',', 'thanks', 'MKM', 'and', '!' |

## 5 RESULT

We have compared the classification accuracy of our method with several other models (Table 4). We make the Dual BERT Embedding Model and Entity Tracking as the baseline. For methods using attention and deep contextualization word representation in some approaches to classify the idiom candidate of cloze-style reading comprehension, some of them use the self-attention for the task. However, they did not perform as well as our model. All models and their variables were trained eight times, making an average for the performance as a result.

From our experiment, as shown as Figure 2, we can get our morel loss reduce with the epochs increase. And we make the cooperation with other two kinds of baselines and our method using the accuracy (Shown as table 4). As we can see, our method have 7.21% higher than the Dual BERT Embedding Model and 14.66% higher than the Entity Tracking in our challenge. That is reasonable that the Dual BERT Embedding Model is designed to solve the challenge of Chinese Idiom Cloze-style Reading comprehension and the Entity Tracking is only designed for the normal Cloze-style

Figure 2: Epoch number and the loss

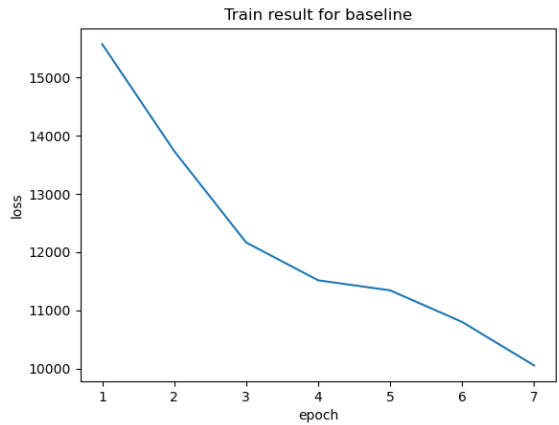

Table 4: Accuracy of task in Idiom Cloze-style Reading comprehension Performance with baselines

| Model | EPIE(%) |
|---|---|
| Tan et al. [17] | 71.02 |
| Our Method | 78.23 |
| Entity Tracking [5] | 63.57 |

Reading comprehension. The Idiom is pretty complex and Idiom always are non-compositional. It is hard to guarantee out methods can work on other kinds of Idiom yet, and from the work of Tan et al, we can see their Dual BERT Embedding Model have a better performance on Chinese Idiom Cloze-style Reading comprehension, they already reach the accuracy of 84.43% in Chinese conditions.

## 6 CONCLUSION AND FUTURE WORK

We developed a new model which carefully performed the Idiom Cloze-style reading comprehension task and made comparisons with common-use algorithms by testing the EPIE dataset. We used different word representation methods and determined that the context details still depend highly on the classification performance. Our method show better performance on the challenge of cloze-style reading comprehension for EPIE dataset. Our method have 7.21% higher than the Dual BERT Embedding Model. With the time limitation, we only work on one dataset for our comprehension. However, it shows up a new milestone for us to solve the challenge of idiom cloze-style reading comprehension.

In our future work, we will explore more attention mechanisms, such as block self-attention [16], or hierarchical attention [21] and hypergraph attention [1]. These approaches can incorporate the information from different representations for the various positions and can capture both local and long-range context dependency. Also, we hope to do more experiments on several idiom dataset, such as CHID [23], LIDIOMS [11]. We want to do more test work and make more stable algorithms to solve the Idiom Cloze-style reading comprehension task with those more kinds of datasets.